\documentclass[nohyperref]{article}

\usepackage{microtype}
\usepackage{graphicx}
\usepackage{subfigure}
\usepackage{booktabs} 

\usepackage{hyperref}

\usepackage[accepted]{icml2022}

\usepackage{amsmath}
\usepackage{amssymb}
\usepackage{mathtools}
\usepackage{amsthm}
\usepackage[symbol,bottom]{footmisc}

\usepackage[capitalize,noabbrev]{cleveref}

\theoremstyle{plain}

\theoremstyle{definition}

\theoremstyle{remark}

\usepackage[textsize=tiny]{todonotes}

\icmltitlerunning{Hamiltonian Domain Translation}

\begin{document}

\onecolumn
\icmltitle{Continuous Methods : Hamiltonian Domain Translation}



\icmlsetsymbol{equal}{*}

\begin{icmlauthorlist}
\icmlauthor{Emmanuel Menier}{1,2}
\icmlauthor{Michele Alessandro Bucci}{1}
\icmlauthor{Mouadh Yagoubi}{2}
\icmlauthor{Lionel Mathelin}{3}
\icmlauthor{Marc Schoenauer}{1}
\end{icmlauthorlist}

\icmlaffiliation{1}{TAU, Inria / LISN, Université Paris-Saclay / CNRS, Orsay, France}
\icmlaffiliation{2}{IRT - SystemX, Palaiseau, France}
\icmlaffiliation{3}{LISN, Université Paris-Saclay \& CNRS, Orsay, France}
\icmlcorrespondingauthor{Emmanuel Menier}{emmanuel.menier@inria.fr}

\icmlkeywords{Deep Learning, Neural ODE, Generative models, invertibility, ICML}

\vskip 0.3in



\printAffiliationsAndNotice{} 

\begin{abstract}
This paper proposes a novel approach to domain translation. Leveraging established parallels between generative models and dynamical systems, we propose a reformulation of the Cycle-GAN architecture. By embedding our model with a Hamiltonian structure, we obtain a continuous, expressive and most importantly invertible generative model for domain translation.
\end{abstract}

\section{Introduction}
\label{introduction}

Domain translation is the process of transforming elements from one domain to another. One can think of applications such as neural style transfer \cite{NST} which is for example used to apply a certain painter's style to photo-realistic images. A common problem encountered in domain translation applications is that, in many cases, paired data is not available during training, which means that the problem has to be formulated in an unsupervised setting. Unsupervised learning is very common in the field of generative modeling, and several architectures have been proposed to deal with the problem of Unsupervised Domain Translation. In this work, we focus on the Cycle-GAN \cite{CycleGan} architecture, which has proved successful in various applications of Unsupervised Domain Translation \footnote{Cycle GAN project page, \url{https://junyanz.github.io/CycleGAN/}}.

Despite its success, the formulation of the Cycle-GAN method has been questioned and shown to be ill-posed. Using results from \citet{Gaussianization}, it can be shown that when considering two distinct domains, there exist an infinity of pairings between the two domains which satisfy the Cycle-GAN objective. This is an issue as the model could get stuck trying to learn wildly inefficient mappings, leading to unsatisfactory optima. This conditioning problem has been explored in depth by \citet{UDT}, as they proposed to use a regularized residual network to learn the mapping between two given domains. Borrowing ideas from optimal transport and dynamical systems, they showed that pushing the training towards simple, low-energy, transformations in latent space leads to learning a sensible and trivially invertible mapping between the two domains of interest. 

The study of the links between dynamical systems theory and deep learning is still to this day a major topic of interest. One can for example cite the identification of residual networks as first order approximations of a time-continuous process which has led to the development of ground-breaking approaches such as neural ordinary differential equations (Neural ODE \cite{NeuralODE}) or invertible neural networks \cite{InvertResNet}. 

Building on this existing connection, as well as the work of de~Bezenac \textit{et al.}, we propose a formulation of unsupervised domain translation as a continuous time process with conservation guarantees which ensure invertibility by construction. The proposed architecture learns the dynamics of the transformation as a Hamiltonian dynamical system. Hamiltonian systems are typically used in General Mechanics to describe the evolution of conservative systems. They preserve a quantity, called the Hamiltonian, along their trajectory. Using neural networks to learn Hamiltonian dynamics is an earlier idea that was proposed in \citet{HNN}. However this work proposes to use them to ensure invertibility of the generative process which is a desirable property to ensure the domain translation problem is well-posed. Learning conservative transformations is in fact critical to other generative modeling approaches, such as normalizing flows \cite{NF}.

\section{Method}\label{sec:method}
\subsection{Invertibility and CycleGAN}
Formally, we can look at the two domains as two separate sets $\mathcal{A},\mathcal{B} \subset \mathbb{R}^d$, where $d$ is the dimension of the space, \textit{i.e.} the pixel space for images, or any latent representation space. The goal of unsupervised domain translation is to learn the forward mapping $F:\mathcal{A}\to\mathcal{B}$ as well as the reverse map $R:\mathcal{B}\to\mathcal{A}$ so that the pair $(F,R)$ generates semantically meaningful samples of each domain. That is to say, the generated samples should be indistinguishable from samples in the target domain, while remaining coherent with their corresponding sample in the original domain. 

CycleGan proposes to enforce these constraints by using a combined loss: $\mathcal{L} = \mathcal{L}_{adv} + \mathcal{L}_{cyc}$. The first term $ \mathcal{L}_{adv} = \mathcal{D}(F(\mathcal{A}),\mathcal{B})$ corresponds to an adversarial loss which measures the distance between the generated samples and the target domain. This term ensures that generated samples are indistinguishable from the target domain. In CycleGAN, $\mathcal{D}$ is implemented using Generative Adversarial Networks \cite{GAN}.

The second term in the loss is called the cyclic loss, $\mathcal{L}_{cyc} = \Vert F \circ R (x_\mathcal{A}) - x_\mathcal{A} \Vert + \Vert R \circ F (x_\mathcal{B}) - x_\mathcal{B} \Vert$. This term promotes transformations $F$ that are invertible and such that $R = F^{-1}$. Intuitively, this pushes the CycleGAN architecture towards learning minimal transformations of the samples, so as to retain a maximum of information from the initial sample and simplify the reconstruction $ R \circ F$. This second term is used to ensure coherence between the translated and initial samples. In addition, learning an invertible (thus bijective) map between the two domains is critical at the conceptual level. Indeed, one sample from a given domain should not map to multiple samples in the target domain as only one sample in the target domain should optimally satisfy the trade-off between coherence with the original sample and similarity with the target domain.

\subsection{Continuous models and Hamiltonian Neural Networks}

The previous paragraph outlined the importance of ensuring the translation map is invertible to relax the learning problem. In fact, this is not specific to the domain translation problem, as invertibility of learned maps has been linked to classical deep learning problems such as vanishing/exploding gradients in recurrent neural networks \cite{RNNConditioning}, or the training of other generative models like normalizing flows \cite{NF}. Several approaches have been proposed to push learned models towards invertibility \cite{SpectralNorm,constrainedNF}, however, they often impose significant constraints on the structure and expressivity of the models, leading to important training costs. 

In this work, we propose to use a natural formulation for invertible transformations. Exploiting the parallel between the residual networks used in numerous image processing approaches, and ordinary differential equations, we propose to define domain translation as a continuous system. Starting at $t=0$ with samples from one domain $x_{t=0} \in \mathcal{A}$, we learn a transport flow $f_\theta$ so that, at $t=T$, $x_{t=T} \in \mathcal{B}$:

\begin{equation}
    \frac{d x}{dt} = f_\theta (x), \quad \mathrm{s.t.} \quad x_0 \in \mathcal{A},
    \: x_T \in \mathcal{B}
\end{equation}

This formulation is not enough to ensure invertibility of the transformation, as the flow $f_\theta$ could be dissipative, or even unstable. To enforce invertibility, we express the flow $f_\theta$ as a conservative operator using Hamiltonian neural networks inspired from \citet{HNN}. To do so, the samples are divided into two vectors of equal length $x = [p,q]$, (we assume $d$ to be even as a modeling choice). In general mechanics, $p$ and $q$ would respectively describe the position and momentum of the studied entities. In our setting, their significance is more abstract and is defined by another function, called the Hamiltonian $\mathcal{H}_\theta(p,q) : \mathbb{R}^{d/2} \times \mathbb{R}^{d/2} \to \mathbb{R}$, which we parameterize using a neural network, hence:

\begin{equation}
f_\theta(x) = 
\left(
\begin{array}{rcr}
    \frac{d p}{dt} & = & -\frac{\partial \mathcal{H}_\theta}{\partial q} \\
    \frac{d q}{dt} & = & \frac{\partial \mathcal{H}_\theta}{\partial p}
\end{array}
\right)
\end{equation}

Using Neural ODEs and automatic differentiation, the function $\mathcal{H}_\theta$ can be trained to satisfy the transport objective, \textit{i.e.} $x_T \in \mathcal{B}$ given $x_0 \in \mathcal{A}$. Moreover, this formulation is invertible by design as it preserves the quantity $\mathcal{H}_\theta (x)$ along its trajectory. We show below that learning the transformation $f_\theta$ with this formulation allows for the generation of semantically correct samples, without using the cyclic loss required in CycleGAN. Thanks to the conservation properties of the flow $f_\theta$, the inverse map is trivially obtained by integrating the flow backward in time:

\begin{equation}
    x_\mathcal{B} = x_\mathcal{A} + \int_0^T f_\theta(x) dt \quad \Longleftrightarrow \quad x_\mathcal{A} = x_\mathcal{B} + \int_T^0 f_\theta(x) dt\label{eq:reverse_trans}
\end{equation}

\section{Results}

\subsection{Generative results}

We apply our Hamiltonian domain translation approach to image generation tasks. As proposed in \citet{UDT}, we train an encoder $E$ and a decoder $G$ to map images from both domains to a latent space of size $d=128$. This is a common approach in many image processing approaches, as the intrinsic dimension of a given image processing problem is often much lower than the dimension of the pixel space. Thus, encoding images to a low-dimensional latent space reduces the domain translation problem complexity, as well as training costs. 

\begin{figure}[ht]
    \centering
    \includegraphics[width=\textwidth]{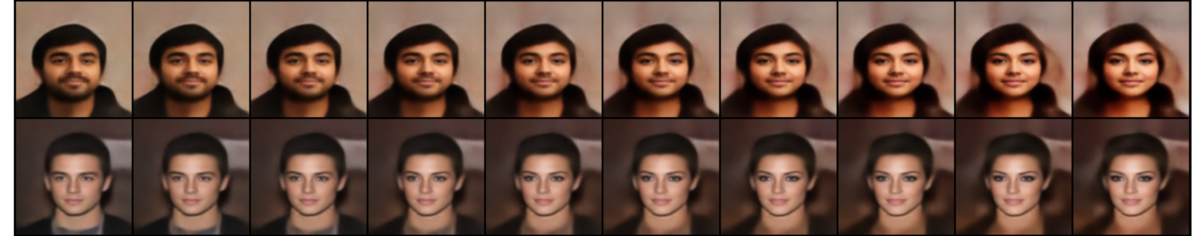}
    \caption{Selected samples of the male to female transport process using the proposed continuous domain translation approach. The transported encodings are decoded at regular time intervals, to illustrate the transformation applied by the model.}
    \label{fig:MTF_samples}
\end{figure}

Once the pair ($E,G$) is trained, it can be used to generate low-dimensional encoded vectors of images of the dataset at hand. We then use our approach to learn the transport flow $f_\theta$. The Hamiltonian $\mathcal{H}$ and discriminator $\mathcal{D}$ are implemented as multi layer perceptron with 3 hidden layers. The continuous flow is learned using the \textit{optimise-then-discretise} version of NeuralODEs. We apply the architecture to the task of translating male samples of the celebA \cite{celebA} dataset to females. Figure \ref{fig:MTF_samples} presents samples generated with this approach. 

\begin{figure}[ht]
    \centering
    \includegraphics[width=\textwidth]{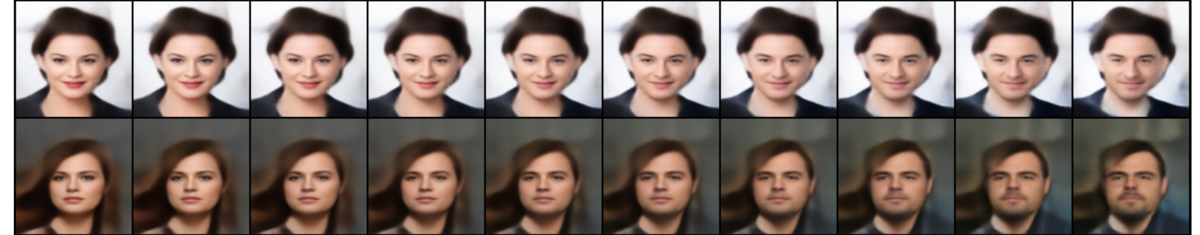}
    \caption{Selected samples of the reverse male to female generation process. NB : These samples are generated using equation~\eqref{eq:reverse_trans} as transport flow was solely trained to map male to females.}
    \label{fig:FTM_samples}
\end{figure}

As shown on Figure \ref{fig:MTF_samples}, decoding the transported samples along the transformation trajectory shows that the flow $f_\theta$ progressively transforms the male samples to females. As expected, the conservative nature of the model promotes transformations that retain non gender-specific features, as we observe that attributes such as pose, skin tone, face shape and background are preserved during the transformation. Figure \ref{fig:FTM_samples} demonstrates an additional interest of the Hamiltonian architecture as we are able to generate males from female samples by simply integrating the flow backward (see Eq.~\ref{eq:reverse_trans}). One should note that these results were obtained without ever training the model to map females to males as we do not compute the cyclic loss used in CycleGAN. These results are similar to the results of \citet{UDT} while no penalization of the magnitude of the flow applied by the model is used but invertibility is enforced instead. 

\begin{figure}[ht]
    \centering
    \includegraphics[width=\textwidth]{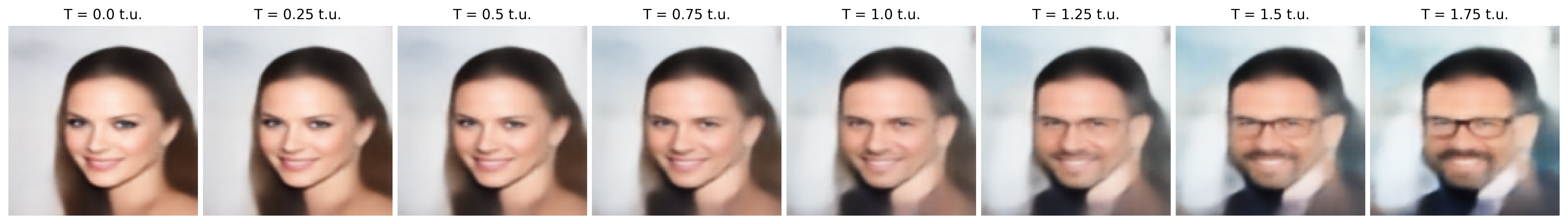}
    \caption{Results of excessive integration of the transport flow. A model trained to map males to females in 1 $t.u.$ is integrated backward for more than twice the map horizon. We observe that the generated samples retain semantic sense for up to about $1.5$ times the training horizon.}
    \label{fig:excessive_integration}
\end{figure}

\subsection{Excessive Integration}

An interesting feature of using a continuous flow to carry out domain translation is that one can gain some insight in the way the model transforms samples. If a flow $f_\theta$ has been trained to map two domains in one time unit (\textit{t.u.}), $T=1$, it can be integrated for a longer period, pushing the transformation further. This is one of the major differences between learning continuous transformations and discrete residual blocks. While residual blocks approximate the flow in specific regions of the latent space, the continuous flow is defined over the whole space. Any trained model starts losing performance once it drifts too far from its training conditions but we observed interesting results when integrating our model for several \textit{t.u.}. Figure \ref{fig:excessive_integration} shows that transported samples retain semantic meaning for up to about one and a half \textit{t.u.}, as the model progressively adds more and more gender-related features such as beards, wider jaws, shorter hair, etc. This generalisation performance can be linked to the conservative architecture of the model which prevents it from diverging to unknown conditions. It also adds to the interest of the approach as it supports the idea that the model is consistent with the structure of the latent space.

\subsection{Training}

It should be noted that, once the autoencoder is trained, learning the flow $f_\theta$ is very inexpensive. The model starts generating semantically coherent samples after a single epoch, and does not require fine tuning between the training of the discriminator $\mathcal{D}$ and the flow $f_\theta$. More formal benchmarking against comparable domain translation methods are planned for the future.

\section{Conclusion}

This work proposes a novel and improved formulation for domain translation. By using a time-continuous approach, we are able to leverage results from general mechanics to obtain a model that is invertible by construction. We show that this model can quickly learn to map two domains of interest, even in a latent space learned prior to training the domain translation architecture.

With the recent success of diffusion models \cite{GLIDE,LDM}, which are based on successive transformations in a pre-defined space, the analogy between generative models and dynamical systems becomes more and more relevant. In this context, the comparison of our proposed method with existing approaches such as normalizing flows and latent diffusion models constitutes the next step to this work, for instance exploring the potential extension of our continuous generative approach to stochastic differential equations \cite{NeuralSDE}.


\bibliography{example_paper}
\bibliographystyle{icml2022}


\end{document}